# A proposal for ethically traceable artificial intelligence

*Christopher A. Tucker, Ph.D., Cartheur Robotics, spol. s r.o., Prague, Czech Republic*

Abstract

Although the problem of a critique of robotic behavior in near-unanimous agreement to human norms seems intractable, a starting point of such an ambition is a framework of the collection of knowledge *a priori* and experience *a posteriori* categorized as a set of synthetical judgments available to the intelligence, translated into computer code. If such a proposal were successful, an algorithm with ethically traceable behavior and cogent equivalence to human cognition is established. This paper will propose the application of Kant's critique of reason to current programming constructs of an autonomous intelligent system.

Introduction

As is oft cited in the literature, a near-universal application of moral imperative in the field of artificial intelligence programming and observed runtime behavior is lacking, theoretical intuitions scattered among five "tribes" [1], existing competitively: Symbolists, evolutionists, Bayesians, analogizers, and connectionists approaching a singularity. A solution is not to satisfy any of them, rather, set criteria that the intuitions themselves approach, as by the definition of the criteria each follows from it instead of the reverse [2].

It can be argued that everything that has been thus far theorized as a means to a solution to the "problem of artificial intelligence" is derived from knowledge *a priori* whose theorems are posited by synthetical judgments. Why this is true can be stated simply: We, as humans, have never experienced nor have been provided judgments of artificial beings save by the process of having developed them, therefore it is impossible to describe them cogently without the specter of fallacious judgments we would never be aware of except for the passage of time. It is, then, for this reason that a new imperative needs to be discovered with which to guide our reason and logic.

I propose the conundrum facing artificial intelligence researchers can be assayed using transcendental logic. In order to clarify the current state of the vexation why a logical imperative cannot be assigned universally because we, as humans, have no experience of how this concept is rooted and therefore its subsequent definition in software is elusive. Admitting this, we must look for philosophical advice and take the viewpoint as an exercise of *pure reason* to find trajectories with which to create universal judgments.

The use of ideological imperative

In searching for a unified framework with which to capture the possible entirety of desirable behaviors for an autonomous intelligent system *a priori*, a powerful model is presented by Immanuel Kant's *Critique of Pure Reason* wherein he notes the universal problem of pure reason. "It is extremely advantageous to be able to bring a number of investigations under the formula of a single problem. For in this manner, we not only facilitate our own labor, inasmuch as we define it clearly to ourselves, but also render it easier for others to decide whether we have done justice in our undertaking. The proper problem of pure reason, then, is contained in the question, 'How are synthetical judgments *a priori* possible?'" [p.12].

Approaching the answer requires the formation of another question regarding the type of knowledge that is desired, e.g., to comprehend the possibility of pure reason in the construction of this science which contains theoretical knowledge *a priori*: How is pure artificial intelligence possible?

Answering this requires an understanding between what we are given about the problem—what really exists—and the natural disposition of the human mind. Does artificial intelligence exist, has it always existed in an apodictic form, and how can the nature of universal human reason find a suitable answer? An answer is yet to be conceived, as the definitions that have been presented in the literature the past half century are inherently contradictory.

A pathway to finding the answer lies in the *Transcendental Doctrine of Elements*, where the particular science, artificial intelligence, is divided under the name of a critique of pure reason. Therefore an "Organon of pure reason would be a compendium of those principles according to which alone all pure cognitions *a priori* can be obtained. The completely extended application of such an organon would afford us a system of pure reason. As this, however, is demanding a great deal, and it is yet doubtful whether any extension of our knowledge be here possible, or if so, in what cases; we can regard a science of the mere criticism of pure reason, its sources and limits, as the propaedeutic to a system of pure reason." [p.15].

The transcendental aesthetic, the first criteria of the doctrine of elements denotes the following: "In whatsoever mode, or by whatsoever means, our knowledge may relate to objects, it is at least quite clear, that the only manner in which it immediately relates to them, is by means of an intuition. But an intuition can take place only in so far as the object is given to us. This, again, is only possible, to man at least, on condition that the object affect the mind in a certain manner." [p.21]. This relates to the inherent sensibility that an object presents to the mind in terms of its representation. As in the case of artificial intelligence, this undetermined object following from an empirical intuition is a phenomenon as it corresponds more to sensation, and can be arranged under certain relations, which constitute its form.

The substance of the form then yields the array of conceptions that have been derived from understanding, to establish the cognition of the object and is properties both anticipated by designers and realized by engineers. "When we call into play a faculty of cognition, different conceptions manifest themselves according to the different circumstances, and make known this faculty, and assemble themselves into a more or less extensive collection, according to the time or penetration that has been applied to the consideration of them. Where this process, conducted as it is, mechanically, so to speak, will end, cannot be determined with certainty. Besides, the conceptions which we discover in the haphazard manner present themselves by no means in order and systematic unity, but are at last coupled together only according to resemblances to each other, and arranged in a series, according to the quantity of their content, from the simpler to the more complex—series which are anything but systematic, through no altogether without a certain kind of method in their construction." [p.56].

This therefore arrives at the construction of an understanding in synthetical judgments. In pure intellectual form, they are presented as:

Table 1. Momenta of thought of synthetical judgments.

| Quantity of judgments | Quality | Relation | Modality |
|---|---|---|---|
| Universal | Affirmative | Categorical | Problematical |
| Particular | Negative | Hypothetical | Assertorical |
| Singular | Infinite | Disjunctive | Apodeictical |

Table 1 contains a list of sets of synthetical judgments divided into momenta categorizations. Kant has noted each corresponds to a property of transcendental logic that now will be disseminated into proposed computer logic. Let us try to formulate a substantive example, which serves to illustrate how this could be done. Assume that the artificial intelligence in question is encapsulated in a single code base such that a developer can work on any part of its program. Let us also assume that this code base consists of styles and patterns of an object-oriented design. In such a form, it is expected the code executes at runtime given conditions of some input, which flows through it forming a pattern or algorithm. This pattern in its entirety is a categorization of synthetical judgment. Because this pattern in known by analysis *a posteriori*, where it has been catalogued and compiled, the execution pathway can be discovered. The next step would be to tag the code along the pathway where it indicates what family, the headers of each column going from left to right of Table 1, and what genus, the columns from top to bottom under each of the headers, it corresponds to.

Once the code is tagged, the tags are gathered to be monitored by an additional process, which is aware of the state of synthetical judgment in the artificial intelligence at any given point in the runtime. It is then theoretically possible to introduce any set of controls on the program that are desirable given behaviors ascribed to each judgment beforehand, tied explicitly to a definition from human experience. In this way, the analogous human experience is no longer separable from logic and therefore allows reason to become an active attribute since the mystery of what is happening within the program at an arbitrarily chosen time is well known.

This forms the basis of universal identity where the perception of phenomenon manifest by the artificial intelligence contains a synthesis of abstract representations of behavior. As such, what Kant calls the *empirical consciousness*, accompanies these different representations to form identity within the subject. Therefore, the problem this paper addresses now takes the relation not as it did before—where the program accompanies every representation with consciousness—but that each representation is joined together sequentially where now the artificial intelligence is conscious of the synthesis of them. Kant states: "Consequently, only because I can connect a variety of given representations in one consciousness, is it possible that I can represent to myself the identity of consciousness in these representations; in other words, the analytical unity of apperception is possible only under the presupposition of a synthetical unity." [p.82].

It is this unity that this paper proposes to add to programming constructs of an artificial intelligence in order that it be ethically traceable and known not only what range of behaviors are programmed into the machine, but those which have the possibility of executing at any point in the runtime.

Discussion and Conclusions

In order to propose a solution for the vexation of the community to derive an algorithm whose behaviors are subject to human norms, a new approach to a solution of artificial intelligence is required. Essentially, in the manner that a programming language is imperative, e.g., using statements that change the program's state—as imperative mood in human languages express commands—thusly an imperative ideology comprised of synthetical judgments can be applied as an intellectual limit in the program's behavior. In this way, a framework with a defined scope is established.

It seems poignantly relevant to introduce this concept now as we are formulating the most basic of laws and regulations for artificial intelligence and autonomous intelligent systems design and manufacture [3]. Rather than trying to sort out the loopholes, confusions, and contradictions, a more efficient approach is to reframe the discussion in terms of fundamental pillars of Western logic and philosophy. In order not to create more confusion or present arguments of a better or worse approach by one philosopher or the other, the work of Immanuel Kant in the *Critique of Pure Reason* is suitable on the foundation that, the solitary question of defining a universal framework whereby to judge artificial intelligence lies in the establishment of a criterion by which it is possible to securely distinguish a pure from an empirical cognition.

The relevance of Kant's philosophy to outline substantive artificial intelligence by a critique of pure reason leaves room for interpretation of implementation but not of the essential framework itself. This is because judgments about what the program *should do* in terms of a system of regulatory axioms and readily traceable to the set of momenta and their empirical manifestation, rather than a random series of trial and error association scenarios. In this way, a list of cogent emotional responses to human interaction is possible given the categorical judgments reactive to a given situation or outcome. It will be for the future to decide whether our intuition can be structured in such a way as to present arguments of objects existing in similitude with our human cognition. Something *must* be proposed in full [4] to address the problems, as we are mired in contradictions about what constitutes the essence of artificial intelligence.